# TENET: Temporal CNN with Attention for Anomaly Detection in Automotive Cyber-Physical Systems

Sooryaa Vignesh Thiruloga
Electrical and Computer Engineering
Colorado State University
Fort Collins CO USA
sooryaa@colostate.edu

Vipin Kumar Kukkala
Electrical and Computer Engineering
Colorado State University
Fort Collins CO USA
vipin.kukkala@colostate.edu

Sudeep Pasricha
Electrical and Computer Engineering
Colorado State University
Fort Collins CO USA
sudeep.pasricha@colostate.edu

*Abstract*— Modern vehicles have multiple electronic control units (ECUs) that are connected together as part of a complex distributed cyber-physical system (CPS). The ever-increasing communication between ECUs and external electronic systems has made these vehicles particularly susceptible to a variety of cyber-attacks. In this work, we present a novel anomaly detection framework called TENET to detect anomalies induced by cyber-attacks on vehicles. TENET uses temporal convolutional neural networks with an integrated attention mechanism to detect anomalous attack patterns. TENET is able to achieve an improvement of 32.70% in False Negative Rate, 19.14% in the Mathews Correlation Coefficient, and 17.25% in the ROC-AUC metric, with 94.62% fewer model parameters, 86.95% decrease in memory footprint, and 48.14% lower inference time when compared to the best performing prior work on automotive anomaly detection.

## I. Introduction

Today's vehicles are becoming increasingly autonomous and connected, to achieve improved safety and fuel efficiency goals. To support this evolution, new technologies such as advanced driver assistance systems (*ADAS*), vehicle-to-vehicle (*V2V*), 5G vehicle-to-infrastructure (*5G V2I*), etc. have emerged [1]. These advances have led to an increase in the complexity of electronic control units (*ECUs*) and the in-vehicle network that connects them. As a result, vehicles today represent distributed cyber-physical systems (CPS) of immense complexity. The ever-growing connectivity to external electronic systems in such vehicles is introducing new challenges, related to the increasing vulnerability of these vehicles to a variety of cyber-attacks [2].

Attackers can use various access points (known as an attack surface) in a vehicle, e.g., Bluetooth and USB ports, telematics systems, and OBD-II ports, to gain unauthorized access to the in-vehicle network. After gaining access to the network, an attacker can inject malicious messages to try and take control of the vehicle. Recent automotive attacks on the in-vehicle network include manipulating speedometer and indicator signals [3], unlocking doors [4], manipulating the fuel level indicator [4], etc. These types of attacks confuse the driver but are not fatal. More complex machine learning-based attacks can cause incorrect traffic sign recognition in a vehicle's camera-connected ECU [5]. In [6], researchers analyzed vulnerabilities in airbag systems and remotely deployed the airbags in a vehicle. These types of attacks can be catastrophic and potentially fatal.

Traditional security mechanism such as firewalls only detect simple attacks and do not have the ability to detect more complex attacks such as those in [5], [6]. With the increasing complexity of vehicular CPS, the attack surface is only going to increase, paving the way for more complex and novel attacks in the future. Thus, there is an urgent need for an advanced attack detection solution that can actively monitor the in-vehicle network and detect complex cyber-attacks. One of the many approaches to achieve this goal is by using an anomaly detection solution (ADS). An ADS can be a hardware or software-based system that continuously monitors the in-vehicle network to detect attacks without any human supervision. Many state-of-the-art ADS use machine learning algorithms to detect cyber-attacks. This is facilitated by the large availability of vehicle network data and more computationally capable ECUs today. At a very high level, the machine learning model in an ADS tries to learn the normal operating behavior of the vehicle system during design and test time. This learned knowledge of the normal system behavior is then used at runtime to continuously monitor for any anomalous behavior, to detect attacks. The major advantage of this approach is that it can detect both known and unknown attacks. Due to its high attack coverage and ability to detect complex attack patterns, we focus on (and make new contributions to) machine learning based ADS for cyber-attack detection in vehicles.

In this work, we propose a novel ADS framework called *TENET* to actively monitor the in-vehicle network and observe for any deviation from the normal behavior to detect cyber-attacks. *TENET* attempts to increase the detection accuracy, receiver operating characteristic (ROC) curve with area under the curve (AUC), Mathews correlation coefficient (MCC) metrics, and minimize false negative rate (FNR) with minimal overhead. Our novel contributions can be summarized as follows:

- We present a temporal convolutional neural attention (TCNA) architecture to learn very-long term temporal dependencies between messages in the in-vehicle network;
- We introduce a metric called divergence score to quantify the deviation from expected behavior;
- We adapt a decision tree-based classifier to detect a variety of attacks at runtime using the proposed metric;
- We compare our *TENET* framework with multiple state-of-the-art ADS frameworks to demonstrate its effectiveness.

## II. Related Work

Several researchers have proposed solutions to detect in-vehicle network attacks. These solutions can be classified as either signature-based or anomaly-based. In this section, we discuss these solutions in detail and present a distinction between the existing works and our proposed *TENET* framework.

The authors in [7] proposed a language-theory based model to derive attack signatures. However, their technique fails to detect attack packets at the initial stages of the attack. In [8], [9] message frequency-based techniques were proposed to detect attacks. A transition matrix-based ADS was proposed in [10] to detect attacks on the controller area network (CAN). However, the approach could not detect complex attacks, such as replay attacks. An entropy-based ADS was presented in [11], [12] to detect in-vehicle network attacks. However, these techniques fail to detect small variations in the entropy and modifications in CAN message data. In [13], the Hamming distance between messages was used to detect attacks. However, this approach incurs a high computational overhead. In [14], ECUs were fingerprinted using their voltage measurements during message transmission and reception. However, as this method is only applied at the physical layer, it cannot detect attacks at the application layer. *In general, signature-based ADS approaches can detect attacks in the network with high accuracy and low false-positive rate. However, obtaining all possible attack signatures and frequently updating them is impractical. Moreover, none of these works provide a holistic solution to detect unknown and complex attack patterns.*

This research is supported by grants from NSF (CNS-2132385)

In contrast, anomaly-based solutions attempt to learn the normal system behavior and observe for any abnormal behavior in the network to detect both known and unknown attacks. In [15], the authors used deep neural networks (DNNs) to extract the low dimensional features of transmitted packets and differentiate between normal and attack-injected packets. The authors in [16] used a recurrent neural network (RNN) to learn the normal behavior of the network and used that information to detect attacks at runtime. Several other works, such as [17]-[20], have proposed long short-term memory (LSTM) based ADS to learn the relationship between messages traversing the in-vehicle networks. However, these models are complex and impose high overheads on the ECU. Moreover, these ADS were not tested on complex attack patterns. A gated recurrent unit (GRU) based autoencoder ADS was proposed to learn the normal system behavior in [21]. However, a static threshold approach was used to classify messages, which is unable to capture non-linear behaviors. In [22], an LSTM based encoder-decoder ADS was proposed with attention to reconstruct input messages. A kernel density estimator (KDE) and k-nearest neighbors (KNN) were further used to detect anomalies. But the approach incurs a high overhead on the ECU. An approach that combined an LSTM with a convolutional neural network (CNN) was proposed in [23] to learn the dependencies between messages in a CAN network. However, the model was trained on a labeled dataset in a supervised manner; due to the large volume of in-vehicle CAN message data, labeling the data is impractical.

All these anomaly-based ADS inspired by machine learning suggest that sequence models with LSTMs and GRUs are popular for detecting attacks on vehicles. *However, the increased vehicular CPS complexity today has resulted in very long-term dependencies between messages exchanged between ECUs that cannot be effectively captured using LSTMs and GRUs.* This is because the current time step output of LSTMs and GRUs is heavily influenced by recent time steps compared to time steps in the distant past, which makes it hard to capture very long-term dependencies. Processing very long sequences also exacerbates the computational and memory overhead of LSTMs and GRUs.

*In summary, none of the existing ADS provides a holistic approach that can efficiently learn very long-term dependencies between in-vehicle network messages with a low memory and computational overhead, and also accurately detect a multitude of simple and complex attacks on the vehicle.* Our proposed *TENET* ADS uses a novel TCNA (temporal CNN with attention) model to overcome these shortcomings of state-of-the-art ADS. The next section describes *TENET* in detail. The comparative performance analysis results are presented in section IV.

## III. *TENET* Framework: Overview

The *TENET* framework consists of three phases: *(i)* data collection and preprocessing, *(ii)* learning, and *(iii)* evaluation. The first phase involves collecting in-vehicle network data from a trusted vehicle and preprocessing the collected data. In the learning phase (*offline*), the preprocessed data is used to train a Temporal Convolutional Neural Attention (TCNA) network in an unsupervised manner to learn the normal behavior of the system. In the evaluation phase (*online*), the trained TCNA network is deployed and used to calculate a divergence score (DS), which is then used by a decision tree-based classifier to detect attacks. The overview of our proposed *TENET* framework is shown in Fig.1.

### A. Data Collection and Preprocessing

This first phase of the *TENET* framework involves collecting in-vehicle network data from a trusted vehicle. During the data collection step, it is crucial to ensure that the in-vehicle network is trusted and free from any malware and that the data is collected over a variety of normal operating conditions. Otherwise, the model may learn an incorrect representation of the normal operation of the vehicle. In this work, we recommended splicing into the vehicle network and directly logging the messages using a standard logger such as Vector GL 1000 [24], as this allows one to record any message traversing the network.

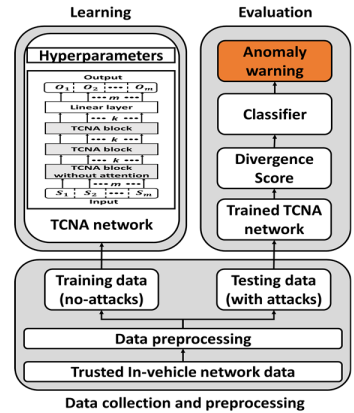

Fig. 1. Overview of the different phases of the *TENET* framework

After data collection, the data is prepared for pre-processing to facilitate easy and efficient training of the machine learning models. Every vehicle message in typical vehicle network protocols (CAN, FlexRay, etc) has a unique identifier (ID) and each message in the dataset is grouped by this unique ID and processed independently. The processed entries are arranged in a tabular format where each row represents a single data sample, and each column represents various unique attributes of the message. Each message has the following attributes (columns): *(i)* unique timestamp corresponding to the log time of the message (used for relative ordering of messages), *(ii)* message ID, *(iii)* number of signals in the message, *(iv)* individual signal values in the message (which together constitute the message payload), and *(v)* label of the message ('0' for no-attack and '1' for attack). Due to the possibility of high variance in message signal values, all signal values of each signal type are scaled between 0 and 1. The learning phase and evaluation phases in *TENET* use training and testing data, respectively. The label values of all samples in the training dataset are set to 0 to represent no-attack data. The test data has a label value of 1 for attack samples and a label value of 0 for no-attack samples. Furthermore, the original training data is split into training (85%) and validation (15%) sets. Details of the training procedure and the model architecture are discussed in the next subsections.

### B. Model Learning

In this subsection, we describe our proposed TCNA network architecture and the training procedure we employed for it. *TENET* uses this TCNA network to learn the normal system behavior of the in-vehicle network in an *unsupervised* manner. The proposed TCNA model takes the sequence of signal values in a message as the input and uses CNNs to predict the signal values of the next message instance, by trying to learn the underlying probability distribution of the normal data.

An early adaptation of CNNs for sequence modeling tasks was presented in [25], where a convolution-based time-delay neural network (TDNN) was proposed for phoneme recognition. To capture very-long term dependencies, traditional CNNs need to employ a very deep network of CNN layers with large filters. Consequently, this increases the number of convolutional operations incurring a high computational overhead. Thus, adapting CNNs directly to sequence modeling tasks in resource constrained automotive systems is not a feasible solution. However, recent advances have enabled the use of CNNs to capture very-long term dependencies with the help of dilated causal convolution (DCC) layers [27]. The *dilation factor* of each DCC layer dictates the number of input samples to be skipped by that layer. The total number of samples influencing the output at a particular time step is called the *receptive field*. Using a larger dilation factor enables an output to represent a wider range of inputs, which helps to learn very-long term dependencies. Unlike RNNs/LSTMs/GRUs, CNNs do not have to wait for the previous time step output to process the input at the current time step. Thus, CNNs can process input sequences in parallel, making them more computationally efficient during both training and testing. Due to these promising properties, we adapt dilated CNNs for learning dependencies between in-vehicle messages in our TCNA model.

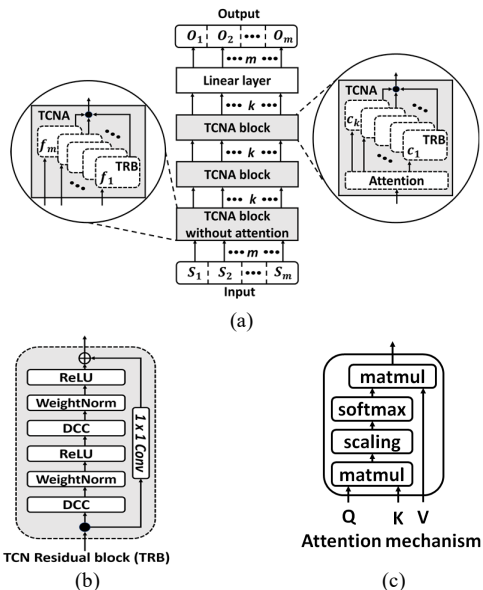

Fig. 2. (a) TCNA network architecture with the internal structure of the TCNA block, (b) TCN residual block showing the various layers of transformation and, (c) the attention mechanism.

We custom designed our TCNA network to consist of three *TCNA blocks*. A TCNA block consists of an attention block and a TCN residual block (TRB), as shown in Fig. 2(a). The input to the first TCNA block is a time series of message data with $n$ signal values as features. This partial sequence from the complete time-series dataset, that is given as the input to the model every time, is called a subsequence. The TRB is inspired by [27] and employs two DCC layers, two weight normalization layers, and two ReLU layers stacked together, as shown in Fig. 2(b). This residual architecture helps to efficiently backpropagate gradients and encourages the reuse of learned features. We enhanced this TRB from [27] by: *(i)* adding an attention layer (discussed later); *(ii)* removing dropout layers to avoid thinning the network and provide our attention block with non-sparse inputs; and *(iii)* avoiding zero-padding the input time-series by computing the length of the subsequence as follows:

$$R = (k-1) * 2^l \qquad (1)$$

where $R$ is the subsequence length, $k$ is the kernel size, and $l$ is the number of DCC layers in the network. This was done because we found that padding zeros to an input sequence distorts the sequential nature of in-vehicle time series data.

The first TCNA block does not contain an attention block, and the inputs are directly fed to the TRB as shown in Fig. 2(a), where, $\{f_1, f_2, ..., f_m\}$ represent multiple channels of the first TRB block, $m$ is equal to the number of features of the inputs, $\{c_1, c_2, ..., c_k\}$ represent multiple channels of the TRB, and $k$ is the number of channels of TRB inputs. The first DCC layer inside the TRB processes each feature of the input sequence as separate *channels* as shown in Fig. 2(a). A 1-D dilated causal convolution operation is performed using a kernel of size two and the number of filters is three times the input features $(m)$ in each DCC layer. The input and output dimensions are the same except for the first TRB. The output from the DCC layer is weight normalized for fast convergence and to avoid explosion of weight values. The weight normalized outputs are passed through a ReLU activation function. This process is repeated one more time inside the TRB. A convolution layer with a filter size of 1×1 is added to make the dimensions of the outputs from the last ReLU activation and the input of the TRB consistent with each other. Each DCC layer in the TRB learns temporal relationships between messages by applying filters to its inputs and updating filter weight values.

Our TCNA block also contains an attention block. Attention mechanisms enable deep neural networks to focus on the important aspects of the input sequences when producing outputs [26]. We devised a scaled dot product attention mechanism and modeled the attention as a mapping of three vectors, namely query *(Q)*, key *(K)*, and value *(V)*. A weight vector is computed by comparing the similarities between the $Q$ and $K$ vectors, and a dot product between the weight vector and the $V$ vector generates the output attention weights. As this attention mechanism does not use the previous output information when generating the attention weights, it is a *self-attention* mechanism. In the context of our proposed TCNA network, self-attention mechanisms can help in identifying important feature maps and enhance the quality of intermediate inputs received by the DCC layers. This further assists with efficient learning of the very-long term dependencies between messages in an in-vehicle network.

The output feature maps of the TRB are given as the input to this attention block, shown in Fig. 2(c). The attention block repeats its inputs to obtain the $Q$, $K$, and $V$ vectors. A scalar-dot product is performed between $Q$ and the transpose of key $(K^T)$ to calculate the similarities between each $Q$ and $K$ vectors. The resultant dot product is scaled by a factor of $1/\sqrt{d_k}$ and passed through a *softmax* layer to calculate attention weights as follows:

$$Attention(Q,K,V) = softmax\left(\frac{QK^T}{\sqrt{d_k}}\right) \cdot V \qquad (2)$$

where the term $d_k$ represents the dimension of the $K$ vector. The attention weights represent the importance of each feature map of the previous DCC layer. The attention weights are then scalar multiplied with $V$ to produce the output of the attention block. Thus, the attention block uses a self-attention mechanism to improve the quality of feature maps that will be received by the subsequent TRBs. Ultimately, as shown in Fig. 2(a), an input sequence flows through the entire TCNA network and is fed to the final linear layer which produces an output of $m$ dimensions. The $m$-dimensional output represents the predicted signal values.

The TCNA network is trained using a rolling window approach. Each window consists of signal values corresponding to the current subsequence. Our TCNA network learns the temporal dependencies between messages inside a subsequence and tries to predict the signal values of the subsequence that are shifted by one time step to the right. We employ a mean squared error (MSE) loss function to compute the prediction error between signal values of the last time step in the predicted subsequence and the last time step of the input subsequence. The error is backpropagated to update the weights for the filters. This process is repeated for each subsequence until the end of the training data, which constitutes one epoch. We train the model for multiple epochs and employ a mini-batch training approach to speed up the training. At the end of each epoch, the model is evaluated using the unseen validation data. An early stopping mechanism is used to prevent model overfitting. The details of the model hyperparameters are discussed later, in section IV-A.

## C. Model Testing

### C.1. Attack model

Here we present details of the various attack scenarios considered in this work. We begin by assuming that the attacker can get access to the in-vehicle network and can modify signal values as well as network parameters at any instance of the vehicle operation. Given this assumption, our *TENET* framework attempts to detect the following complex and most widely seen attack scenarios in the in-vehicle network:

*1) Plateau attack:* This is an attack scenario where the attacker sets a constant value for a signal or multiple signals over the attack interval. It is hard to detect this attack especially when the set constant value is close to the true signal value.

*2) Suppress attack:* In this attack, the attacker tries to suppress a signal value by either disabling the ECU or deactivating the communication controller, effectively resulting in no message being transmitted. It is hard to detect short bursts of these attacks as they could be confused for a missing or delayed message.

*3) Continuous attack:* This attack represents a scenario where the attacker gradually overwrites the true signal value. The attacker then eventually will achieve the target value without triggering most ADS frameworks. These attacks are hard to detect and require an advanced ADS.

*4) Playback attack:* This attack involves the attacker using the previously observed sequences of signal values and trying to replay them again at a later time to trick the ADS. If the ADS is not trained to understand patterns in the sequence of transmitted messages, it will fail to detect these types of advanced attacks.

### C.2. Evaluation phase

We use the trained TCNA network in conjunction with a detection classifier to detect attacks on vehicles at runtime. The high frequency of messages in the in-vehicle network requires a detection classifier that is lightweight and can classify messages quickly with high detection accuracy and minimal overhead. Hence, we use the well-studied categorical variable decision tree-based classifier to detect between normal and attack samples (binary classification) due to their simpler nature, speed, and precise classification capabilities.

A decision tree starts with a single node (*root* node), which branches into possible outcomes. Each of those outcomes leads to additional nodes called *branch nodes*. Each branch node branches off into other possibilities and ends in a *leaf node* giving it a treelike structure. During training, the decision creates the tree structure by determining the set of rules in each branch node based on its input. During testing, the decision tree takes the input and traverses the tree structure until it reaches a leaf node. The evaluation phase begins by splitting the test data with attacks into two parts: *(i)* calibration data, and *(ii)* evaluation data. In the first part, only the calibration data is fed to the trained TCNA network to generate the predicted sequences. We then compute a divergence score (DS) for each signal in every message:

$$DS_i^m(t) = \left(\hat{S}_i^m(t) - S_i^m(t+1)\right) \forall\ i \in [1, N_m], m \in [1, M] \quad (3)$$

where *m* represents the $m^{th}$ message sample and *M* represents the total number of message samples, *i* represents the $i^{th}$ signal of the $m^{th}$ message sample and $N_m$ represents the total number of signals in the $m^{th}$ message, *t* represents the current time step, $\hat{S}_i^m(t)$ represents the $i^{th}$ predicted signal value of the $m^{th}$ message at time step *t*, and $S_i^m(t+1)$ represents the true $i^{th}$ signal value of the $m^{th}$ message sample at time step *t + 1*.

The DS is higher during an attack as the TCNA model is trained on the no-attack data and fails to predict the signal values correctly in the event of an attack. This sensitive nature of the DS to attacks makes it a good candidate for the input to our detection classifier. Moreover, the group of signal level DS for each message sample is stacked together to obtain a DS vector. We train the decision tree classifier using this DS vector as input to learn the distribution of both no-attack samples and attack samples. We use the unseen evaluation data (that has both attack and no-attack samples) to evaluate the performance of *TENET*.

## IV. Experimental Setup

To evaluate the effectiveness of the *TENET* framework, we conducted various experiments. We compared *TENET* against three state-of-the-art prior works on ADS: RN [16], INDRA [21], and HAbAD [22]. Together, these approaches reflect a wide range of sequence modeling architectures. RN [16] uses RNNs to increase the dimensionality of input signal values and reconstruct the input signal at the output by minimizing MSE. The trained RN model scans continuously for large reconstruction errors at runtime to detect anomalies over in-vehicle networks. INDRA [21] uses a GRU-based autoencoder that reconstructs input sequences at the output by reducing the MSE reconstruction loss. At runtime, INDRA utilizes a pre-computed static threshold to flag anomalous messages. HAbAD [22] uses an LSTM based autoencoder model with attention to detect anomalies in real-time embedded systems. HAbAD uses a supervised learning detector that combines a kernel density estimator (KDE) and k-nearest neighbors (KNN) algorithm to detect anomalies. The comparisons of *TENET* with the above-mentioned ADS are presented in subsections IV-B and IV-C.

We adopted an open-source CAN message dataset developed by ETAS and Robert Bosch GmbH [17] to train our model, and the comparison works. The dataset consists of multiple CAN messages with different number of signals that were modeled after real-world vehicular network information. Moreover, the dataset has a distinct training set that has normal CAN messages and a labeled testing dataset for different types of attacks. For training and validation, we used the training dataset from [17] without any attack scenarios in an unsupervised manner. We tested our proposed *TENET* framework, and all comparison works by modeling various real-world attacks (discussed in section III-C.1) using the test dataset in [17]. Note that *TENET* can be easily adapted to other in-vehicle network protocols such as Flexray and Ethernet, as it relies only on the message payload information. However, the lack of any openly available datasets for these protocols prevents us from showing results on them.

We used PyTorch 1.8 to model and train various machine learning models including *TENET*, and the comparison works. Our framework uses 85% of data for training and the remaining 15% for validation. We trained *TENET* for 200 epochs with an early stopping mechanism that constantly monitors the validation loss after each epoch. If no improvement in validation loss is observed in the past 10 (patience) epochs, training is terminated. We used MSE to compute the prediction error and the ADAM optimizer with a learning rate of 1e-4. We employed a rolling window approach (discussed in Section III-B) with a batch size of 256, and a subsequence length of 64. We used scikit-learn to implement the decision tree classifier, with the *gini* criterion, and *best* splitter to detect anomalies based on the divergence score.

Before discussing the results, we define performance metrics that we used in the context of ADS. We classify a message as a true positive *(TP)* only if the model detects a true attack as an anomaly, and a true negative *(TN)* is when a normal message is detected as a no-attack message. When the model detects a normal message as an anomalous message it is defined as a false positive *(FP)*, whereas an actual anomalous message which is not detected is a false negative *(FN)*. Using these definitions, we evaluate the ADS based on <u>four</u> different performance metrics:

*(i) Detection Accuracy*: which quantifies the ability of the ADS to detect an anomaly correctly, as defined below:

$$Detection\ accuracy = \frac{TP+TN}{TP+FP+TN+FN} \quad (4)$$

*(ii) Receiver Operating Characteristic (ROC) curve with area under the curve (AUC)*: which measures the ADS performance as the area under the curve in a plot between the true positive rate (TPR) and false positive rate (FPR):

$$TPR = \frac{TP}{TP+FN} \qquad FPR = \frac{FP}{FP+TN} \quad (5)$$

*(iii) False Negative Rate (FNR)*: which quantifies the probability that a *TP* will be missed by the ADS (lower is better):

$$FNR = \frac{FN}{FN+TP} \quad (6)$$

*(iv) Mathews Correlation Coefficient (MCC)*: which provides an accurate evaluation of the ADS performance while working with imbalanced datasets, as defined below:

$$MCC = \frac{(TP*TN)-(FP*FN)}{\sqrt{(TP+FP)(TP+FN)(TN+FP)(TN+FN)}} \quad (7)$$

Another metric that is sometimes used is the F-1 score, which is the harmonic mean of precision and recall. As both precision and recall do not include the true negatives in their computation, the F-1 score metric fails to represent the true performance of the classifier. Unlike the F-1 score metric, the MCC metric that we consider includes all the cells of the confusion matrix, thus providing a much more accurate evaluation of the frameworks.

### A. Receptive Field Length Sensitivity Analysis

In the first experiment, we compare the performance of our TCNA architecture with four different receptive field lengths while the remaining hyperparameters are unchanged. We conduct this analysis to evaluate whether very long receptive lengths can help with a better understanding of normal system behavior. All the variants are evaluated based on their performance on two training metrics: average training loss and average validation loss, and the best model is selected for further comparisons. The average training loss value represents the average loss between the predicted behavior and observed behavior of each iteration in the training data. In contrast, the average validation loss represents the average loss between the predicted behavior and the observed behavior of each iteration in the validation data.

Table I: TCNA variants with different receptive field lengths

|  | Receptive field lengths | | | |
|---|---|---|---|---|
|  | 16 | 32 | 64 | 128 |
| Average training loss | 4.1e-4 | 3e-4 | **2.5e-4** | 6.8e-4 |
| Average validation loss | 5.5e-4 | 4.3e-4 | **2.9e-4** | 9.3e-4 |

Table I shows the average training and validation loss of the four variants of TCNA. We can observe that a receptive length of 64 has the lowest average training and validation loss. Therefore, we select 64 as our receptive field value, which is twice the maximum receptive field length presented in one of the comparison works (sequence length of 32 in [22]). This long receptive field length enables us to more effectively learn very long-term dependencies in the input time series data and allows us to better understand the normal vehicle operating behavior.

### B. Prior Work Comparison

In this subsection, we compare our *TENET* framework with the state-of-the-art ADS works RN [16], INDRA [21], and HAbAD [22]. The results of the comparison on the metrics discussed in the previous section are as shown in Fig. 3. From Fig. 3(a)-(d), *TENET* outperforms all comparison works for all four metrics under various attack scenarios.

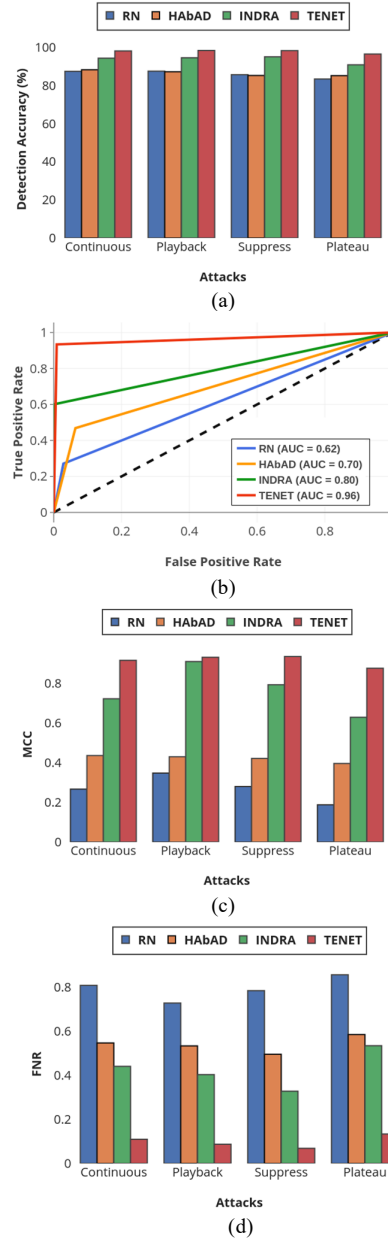

Fig. 3. Comparison of (a) detection accuracy, (b) ROC-AUC for playback attack, (c) MCC, and (d) FNR for *TENET* and ADS from prior work.

Table II: Relative % improvement of *TENET* vs. other ADS

| Prior ADS Works | Detection accuracy | ROC-AUC | MCC | FNR |
|---|---|---|---|---|
| INDRA [21] | 3.32 | 17.25 | 19.14 | 32.70 |
| HABAD [22] | 9.07 | 26.50 | 49.26 | 44.05 |
| RN [16] | 9.48 | 37.25 | 64.3 | 69.47 |

Table II summarizes the average relative percentage improvement of *TENET* over the comparison works for all attack scenarios. Compared to the best performing prior work (INDRA),

*TENET* achieves an improvement of 3.32% in detection accuracy, 17.25% in ROC-AUC for playback attacks (we only show a playback attack for representing the ROC-AUC as it is the most difficult attack to detect), 19.14% in MCC, and 32.70% in FNR.

*In summary*, our *TENET* framework with a customized TCNA network outperforms all prior recurrent architectures with and without attention, due to its ability to capture very-long term dependencies in time-series data. Moreover, the attention mechanism within the TCNA improves the quality of the outputs of the TRB enabling efficient learning of very-long term dependencies. Thus, our TCNA network with the decision tree classifier represents a formidable anomaly detection framework.

### C. Memory Overhead and Latency Analysis

Lastly, we compare the number of trainable parameters, the memory footprint, and inference time of the *TENET* framework, and the comparison ADS works to evaluate their memory and computational overheads. Table III shows the memory footprint, model parameters, and average inference latency of *TENET* and the other ADS. It is important to consider the memory and latency overhead of ADS models because automotive ECUs are resource constrained and it is crucial to have an ADS that does not interfere with the normal operation of safety-critical automotive applications. All results are obtained for deployment on an NVIDIA Jetson TX2 with dual-core ARM Cortex-A57 CPUs, which has specifications similar to real-world ECUs.

Table III: Memory, model size, and inference latency analysis

| ADS Framework | Memory footprint (KB) | Model parameters | Inference time ($\mu s$) |
|---|---|---|---|
| TENET | 59.62 | 6064 | 250.24 |
| RN [16] | 7.2 | 1300 | 412.50 |
| INDRA [21] | 453.8 | 112900 | 482.10 |
| HAbAD [22] | 261.63 | 64484 | 1370.10 |

It can be observed that *TENET* has the second lowest number of model parameters and memory footprint over all the other comparison works except RN [16]. Even though RN has the least number of model parameters and memory footprint, it fails to effectively capture the temporal dependencies between messages, resulting in very poor performance, as can be seen in Fig. 3(a)-(d). Compared to INDRA and HAbAD, *TENET* achieves a reduction of 86.86% and 77.21% in memory footprint, and a reduction of 94.62% and 90.59% in the number of trainable model parameters. *TENET* is able to achieve high performance with significantly fewer trainable parameters because of the fewer filters used by each DCC layer in the TCNA network. This is achieved using the attention block in TCNA which improves the quality of the outputs of each TRB thus eliminating the need for more filters. Moreover, *TENET* also has the lowest inference time with an average of 56.43% reduction against all comparison works. *TENET* is able to achieve faster inferencing because, unlike recurrent architectures, *TENET* employs CNNs to process multiple subsequences in parallel, which helps reduce the inference time. Thus, *TENET* is able to achieve superior performance across various attack scenarios in automotive platforms with minimal memory and computational overhead.

### VI. Conclusion

In this paper, we have proposed a novel anomaly detection framework called *TENET* for automotive cyber-physical systems based on Temporal Convolutional Neural Attention (TCNA) networks. We also proposed a metric called the divergence score (DS), which measures the deviation of the predicted signal value from the actual signal value. We compared our framework with the best-known prior works that employ a variety of sequence model architectures for anomaly detection. Compared to the best performing prior work, *TENET* achieves an improvement of 3.32% in detection accuracy, 17.25% in ROC-AUC, 19.14% in MCC, and 32.70% in FNR metric with 94.62% fewer model parameters, 86.95% decrease in memory footprint, and 48.14% lower inference time. Given the proliferation of connected vehicles with large attack surfaces on the roads today, the promising results in this work highlight a compelling potential for deploying *TENET* to achieve fast, low-footprint, and accurate anomaly detection in emerging automotive platforms.